\documentclass{article}

\usepackage{PRIMEarxiv}

\usepackage[utf8]{inputenc} 
\usepackage[T1]{fontenc}    
\usepackage{hyperref}       
\usepackage{url}            
\usepackage{booktabs}       
\usepackage{amsfonts}       
\usepackage{amsmath}        
\usepackage{nicefrac}       
\usepackage{microtype}      
\usepackage{fancyhdr}       
\usepackage{graphicx}       
\usepackage{float}
\usepackage{caption}
\usepackage{algorithmic}
\usepackage{algorithm}

\title{Zero-Shot Artifact2Artifact: Self-incentive artifact removal for photoacoustic imaging without any data}

\author{
  Shuang Li \\
  College of Future Technology \\
  Peking University \\
  Beijing, China \\
  \texttt{jaeger\_ls@stu.pku.edu.cn} \\
  \And
  Qian Chen \\
  College of Future Technology \\
  Peking University \\
  Beijing, China \\
  \texttt{chen\_qian@stu.pku.edu.cn} \\
  \And
  Chulhong Kim \\
  Department of Electrical Engineering, Convergence IT Engineering, Mechanical Engineering, \\
  and Medical Science and Engineering, Medical Device Innovation Center, \\
  Pohang University of Science and Technology (POSTECH), \\
  77 Cheongam-ro, Nam-gu, Pohang, Gyeongbuk 37673, Republic of Korea \\
  \texttt{chulhong@postech.edu} \\
  \And
  Seongwook Choi \\
  Department of Electrical Engineering, Convergence IT Engineering, Mechanical Engineering, \\
  and Medical Science and Engineering, Medical Device Innovation Center, \\
  Pohang University of Science and Technology (POSTECH), \\
  77 Cheongam-ro, Nam-gu, Pohang, Gyeongbuk 37673, Republic of Korea \\
  \texttt{swchoi715@postech.ac.kr} \\
  \And
  Yibing Wang \\
  College of Future Technology \\
  Peking University \\
  Beijing, China \\
  \texttt{ddffwyb@pku.edu.cn} \\
  \And
  Yu Zhang \\
  College of Future Technology \\
  Peking University \\
  Beijing, China \\
  \texttt{zyuaiyi1\_@stu.pku.edu.cn} \\
  \And
  Changhui Li* \\
  College of Future Technology \\
  Peking University \\
  Beijing, China \\
  \texttt{chli@pku.edu.cn} \\
}

\begin{document}

\maketitle

\newcommand\nnfootnote[1]{%
  \begin{NoHyper}
  \renewcommand\thefootnote{}\footnote{#1}%
  \addtocounter{footnote}{-1}%
  \end{NoHyper}
}
\nnfootnote{* Corresponding author.}

\begin{abstract}
Photoacoustic imaging (PAI) uniquely combines optical contrast with the penetration depth of ultrasound, making it critical for clinical applications. However, the quality of 3D PAI is often degraded due to reconstruction artifacts caused by the sparse and angle-limited configuration of detector arrays. Existing iterative or deep learning-based methods are either time-consuming or require large training datasets, significantly limiting their practical application. Here, we propose Zero-Shot Artifact2Artifact (ZS-A2A), a zero-shot self-supervised artifact removal method based on a super-lightweight network, which leverages the fact that reconstruction artifacts are sensitive to irregularities caused by data loss. By introducing random perturbations to the acquired PA data, it spontaneously generates subset data, which in turn stimulates the network to learn the artifact patterns in the reconstruction results, thus enabling zero-shot artifact removal. This approach requires neither training data nor prior knowledge of the artifacts, and is capable of artifact removal for 3D PAI. For maximum amplitude projection (MAP) images or slice images in 3D PAI acquired with arbitrarily sparse or angle-limited detector arrays, ZS-A2A employs a self-incentive strategy to complete artifact removal and improves the Contrast-to-Noise Ratio (CNR). We validated ZS-A2A in both simulation study and \( in\ vivo \) animal experiments. Results demonstrate that ZS-A2A achieves state-of-the-art (SOTA) performance compared to existing zero-shot methods, and for the \( in\ vivo \) rat liver, ZS-A2A improves CNR from 17.48 to 43.46 in just 8 seconds. The project for ZS-A2A will be available in the following GitHub repository: \href{https://github.com/JaegerCQ/ZS-A2A}{https://github.com/JaegerCQ/ZS-A2A}.

\end{abstract}

\keywords{3D photoacoustic imaging \and artifact removal \and zero-shot \and sparse or angle-limited detector array}

\section{Introduction}

Photoacoustic imaging (PAI) combines ultrasound detection with optical absorption contrast, establishing itself as a powerful, label-free optical imaging modality in clinical applications, which enables high spatial resolution imaging of living tissue at depths of several centimeters~\cite{park2024clinical, wang2012photoacoustic, assi2023review, dean2017advanced, lin2022emerging, ntziachristos2024addressing}. Furthermore, PAI supports three-dimensional (3D) imaging of biological tissues, and recent advancements in spherical and planar array technologies for 3D imaging \cite{matsumoto2018label, matsumoto2018visualising, ivankovic2019real, dean2013portable, nagae2018real, kim2023wide, piras2009photoacoustic, heijblom2012visualizing} have significantly contributed to the progress of 3D PAI. However, due to cost and technical difficulties, most real-time 3D PAI systems tends to have sparse and angle-limited sensor arrays in reality, which desires advanced algorithms to reduce artifacts and enhance Contrast-to-Noise Ratio (CNR) of PAI results reconstructed with universal back-projection (UBP) algorithm. 

To address this problem caused by sparse and angle-limited sensor arrays, researchers have employed both iterative reconstruction (IR) and deep-learning-based methods to improve image quality. For 3D PAI reconstruction, iterative methods often suffer from extremely high memory consumption and long computation time~\cite{paltauf2002iterative, wang2012investigation, huang2013full, zhu2023mitigating, arridge2016adjoint, shang2019sparsity}. Even relatively advanced iterative methods~\cite{li2024sliding} can take hours to reconstruct a volume (e.g., a $25.6 \text{ mm} \times 25.6 \text{ mm} \times 25.6 \text{ mm}$ region at $0.1 \text{ mm}$ resolution) of the PA data acquired by a common photoacoustic hemispherical system. Additionally, current artifact removal and image quality enhancement methods based on deep learning—including both supervised~\cite{allman2018photoacoustic, deng2020unet, guan2019fully, zhang2020new} and unsupervised~\cite{lu2021artifact, zhong2024unsupervised} approaches—typically require extensive datasets for network pretraining. These methods show poor transferability, which is nearly impractical for 3D PAI systems with significant system variability.

Thus, a self-supervised neural network-based method that does not require any pretraining on external datasets is more suitable. It can efficiently remove artifacts in 3D reconstruction results within a very short time. Reconstruction artifacts in PAI share certain similarities with noise in images—their spatial distribution and fluctuations are significantly different from true signals. To achieve zero-shot denoising, a method called Zero-Shot Noise2Noise~\cite{mansour2023zero}, which builds upon Noise2Noise~\cite{lehtinen2018noise2noise} and Neighbor2Neighbor~\cite{huang2021neighbor2neighbor}, generates two down-sampled images from noisy input data through fixed filtering and optimizes the network using residual and consistency losses, enabling the network to learn noise patterns from the noisy images themselves.

Inspired by Zero-Shot Noise2Noise, we propose Zero-Shot Artifact2Artifact (ZS-A2A), a zero-shot self-supervised artifact removal method based on a super-lightweight network. Unlike Zero-Shot Noise2Noise, ZS-A2A is tailored to the physical model of reconstruction artifacts in PAI. It generates pairs of down-sampled reconstruction results by applying randomized perturbations to the input data and further optimizes the network with residual and consistency losses. This enables the neural network to learn the artifact patterns within the input data, facilitating super-efficient artifact removal for PAI reconstruction results.

ZS-A2A does not require any pretraining on external datasets or any prior knowledge about the artifacts. It can complete artifact learning and inference for a $256 \times 256$ input slice in about 8 seconds, and for a $256 \times 256 \times 256$ volume of 3D PAI in about 25 minutes. We validated the artifact removal capability of ZS-A2A on both simulated data and real \( in\ vivo \) rat liver and kidney data.

\section{Method}

\subsection{Randomized perturbations-based artifact patterns learning for photoacoustic imaging}

The general expression for the photoacoustic pressure in a non-viscous, infinitely large homogeneous medium can be written as~\cite{xu2005universal}:

\begin{equation}
\left(\nabla^2 - \frac{1}{c^2} \frac{\partial^2}{\partial t^2} \right) p(\vec{r}, t) = -p_0(\vec{r}) \frac{d \delta(t)}{d t},
    \label{eq:eq1}
\end{equation}

where $\mathit{c}$ is the velocity of the acoustic wave in the medium, $\mathit{\delta(t)}$ is the pulsed electromagnetic (EM) waves, and the acoustic wave $p(\vec{r}, t)$ at position $\vec{r}$ and time $\mathit{t}$ is triggered by the initial source $p_0(\vec{r})$.

Using Gauss's theorem and the properties of the Green's function, the universal back-projection (UBP) formula can be deduced as~\cite{xu2005universal}:

\begin{equation}
    p_0^{(b)}(\vec{r}) = \int_{\Omega_0} b\left(\vec{r_0}, \overline{t} = \left|\vec{r} - \vec{r_0}\right|\right) \frac{d\Omega_0}{\Omega_0},
    \label{eq:eq2}
\end{equation}

where $\mathit{\overline{t} = c t}$, $b\left(\vec{r_0}, \overline{t}\right) = 2p\left(\vec{r_0}, \overline{t}\right) - 2\overline{t} \frac{\partial p\left(\vec{r_0}, \overline{t}\right)}{\partial \overline{t}}$ represents the back-projection term related to the measurement at position $\vec{r_0}$, $d\Omega_0 = \frac{dS_0}{\left|\vec{r} - \vec{r_0}\right|^2} \cdot \frac{n_0^s \cdot \left(\vec{r} - \vec{r_0}\right)}{\left|\vec{r} - \vec{r_0}\right|}$ is the solid angle subtended by detector element $\mathit{dS_0}$ relative to the reconstruction point $\mathit{P}$~\cite{xu2005universal}.

\begin{figure}[ht]
    \centering
    \includegraphics[width=1.0\linewidth]{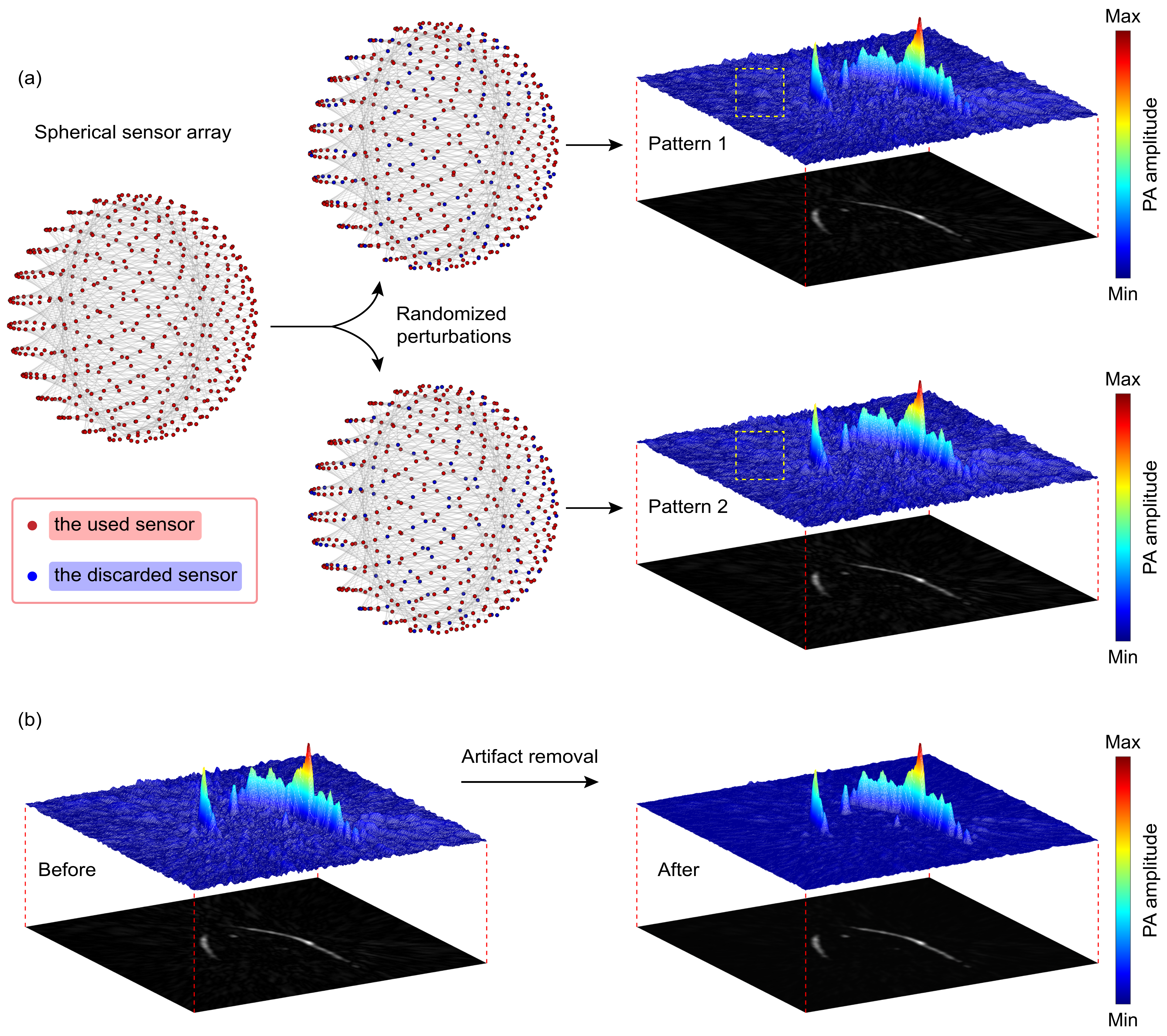}
    \caption{Randomized perturbations-based artifact patterns learning. (a) Applying randomized perturbations to reconstruction results by discarding data from randomly selected detectors. (b) Artifact removal based on the learned artifact patterns from randomized perturbations.}
    \label{fig:fig1}
\end{figure}

When the detection surface consists of a sparsely distributed spherical array, due to the reliance of the UBP algorithm (Eq.~\ref{eq:eq2}) derivation on Gauss's theorem, the condition for a continuous integration surface cannot be fully satisfied. As a result, reconstruction artifacts arise and become more pronounced as the degree of sparsity and the extent of missing angular coverage in the detection surface increase, eventually overwhelming the reconstructed true signal.

In general, the reconstructed true PA signal and the artifacts exhibit distinctly different fluctuation characteristics. To exploit this difference, during the reconstruction process, we take the existing detector signals n (from 1 to N, where N is the total number of detectors) and randomly select M detectors to form subsets of data (Eq.~\ref{eq:eq3}). These subsets are then used for separate PAI reconstructions. By selectively discarding certain detector signals to generate randomized subsets of detector data, we effectively introduce random perturbations into the reconstruction process based on the physical model of PAI (Fig.~\ref{fig:fig1}(a)).

\begin{equation}
\Gamma = \{n_1, n_2, \dots, n_M \mid n_i \in \{1, 2, \dots, N\}, n_i \text{ is randomly selected, and } M \leq N\},
    \label{eq:eq3}
\end{equation}

where \( \Gamma \) represents the subset of detector indices that are randomly selected.

\begin{figure}[ht]
    \centering
    \includegraphics[width=1.0\linewidth]{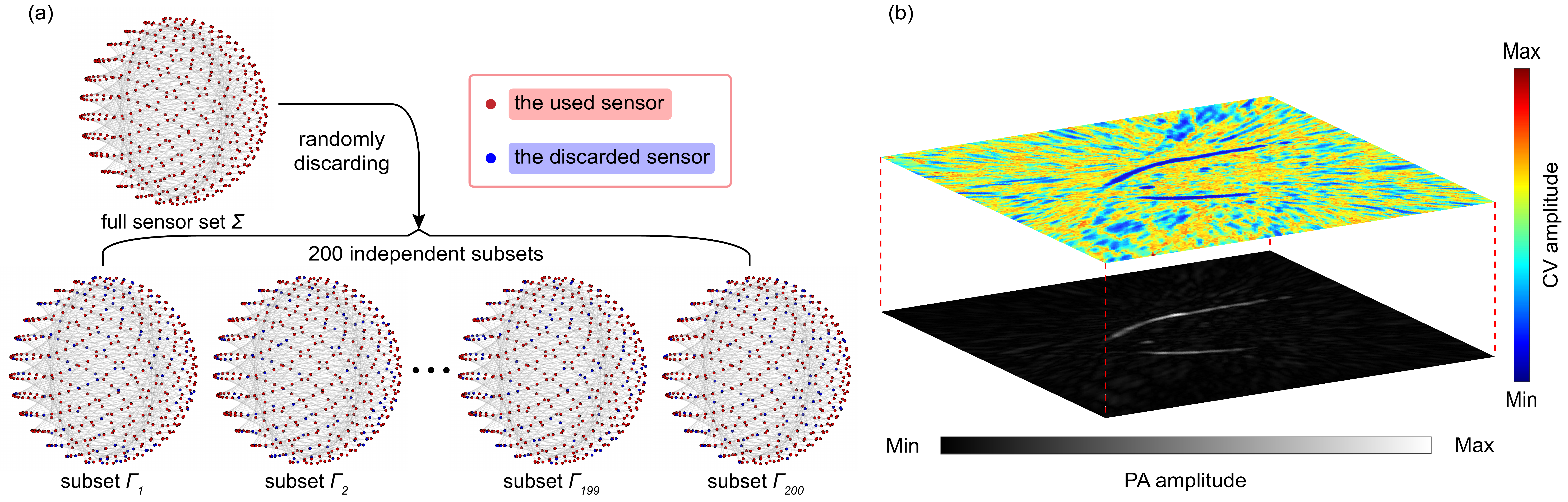}
    \caption{Numerical instability of artifact fluctuations under random perturbations. (a) Subset sensor data generated from 200 independent random discarding perturbations. (b) Coefficient of variation (CV) of artifacts and PA signals in the reconstructed image.}
    \label{fig:fig2}
\end{figure}

The fluctuation characteristics of artifacts and true PA signals can be reflected by the coefficient of variation (CV) of the reconstructed values at locations of spatially reconstructed PA signals and potential artifact positions. The coefficient of variation is a statistical measure of data dispersion, representing the degree of fluctuation in data relative to its mean. Its calculation formula is shown in Eq.~\ref{eq:eq4}. Generally, the larger the coefficient of variation, the greater the fluctuation and the lower the stability of the data; conversely, the smaller the coefficient of variation, the less the fluctuation and the better the stability of the data. Using a random discarding strategy, 200 subsets of detector data were generated (Fig.~\ref{fig:fig2}(a)). Based on independent reconstruction results for these 200 subsets, the coefficient of variation for the reconstructed values under perturbation of the detector data was analyzed (Fig.~\ref{fig:fig2}(b)). It is observed that the artifact regions exhibit very high coefficients of variation, indicating extreme sensitivity to changes in detector data, whereas the PA signal regions have much smaller coefficients of variation, demonstrating strong stability and minimal fluctuation.

\begin{equation}
    CV = \frac{\sigma}{\mu} \times 100\%,
    \label{eq:eq4}
\end{equation}

where \( \sigma \) represents the standard deviation of the reconstructed point's value, and \( \mu \) represents the mean of the reconstructed point's value.

In PAI with non-extremely sparse detector distributions, the redundancy in the recorded data allows the reconstruction to adapt to certain missing information without losing the overall integrity of the PA source. Thus the randomized subset strategy leverages the unique spatial distribution properties of reconstruction artifacts versus true PA signals, helps the model learn the artifact pattern and further remove artifacts in the full-sampled reconstruction result (Fig.~\ref{fig:fig1}(b)). 

\subsection{Zero-Shot Artifact2Artifact}

Based on the fact that reconstruction artifacts are sensitive to irregularities caused by unordered data loss, we further develop a lightweight neural network to establish a zero-shot self-supervised framework, Zero-Shot Artifact2Artifact (ZS-A2A), to remove artifacts from the reconstruction results of photoacoustic imaging and enhance image quality. 

\begin{figure}[ht]
    \centering
    \includegraphics[width=0.9\linewidth]{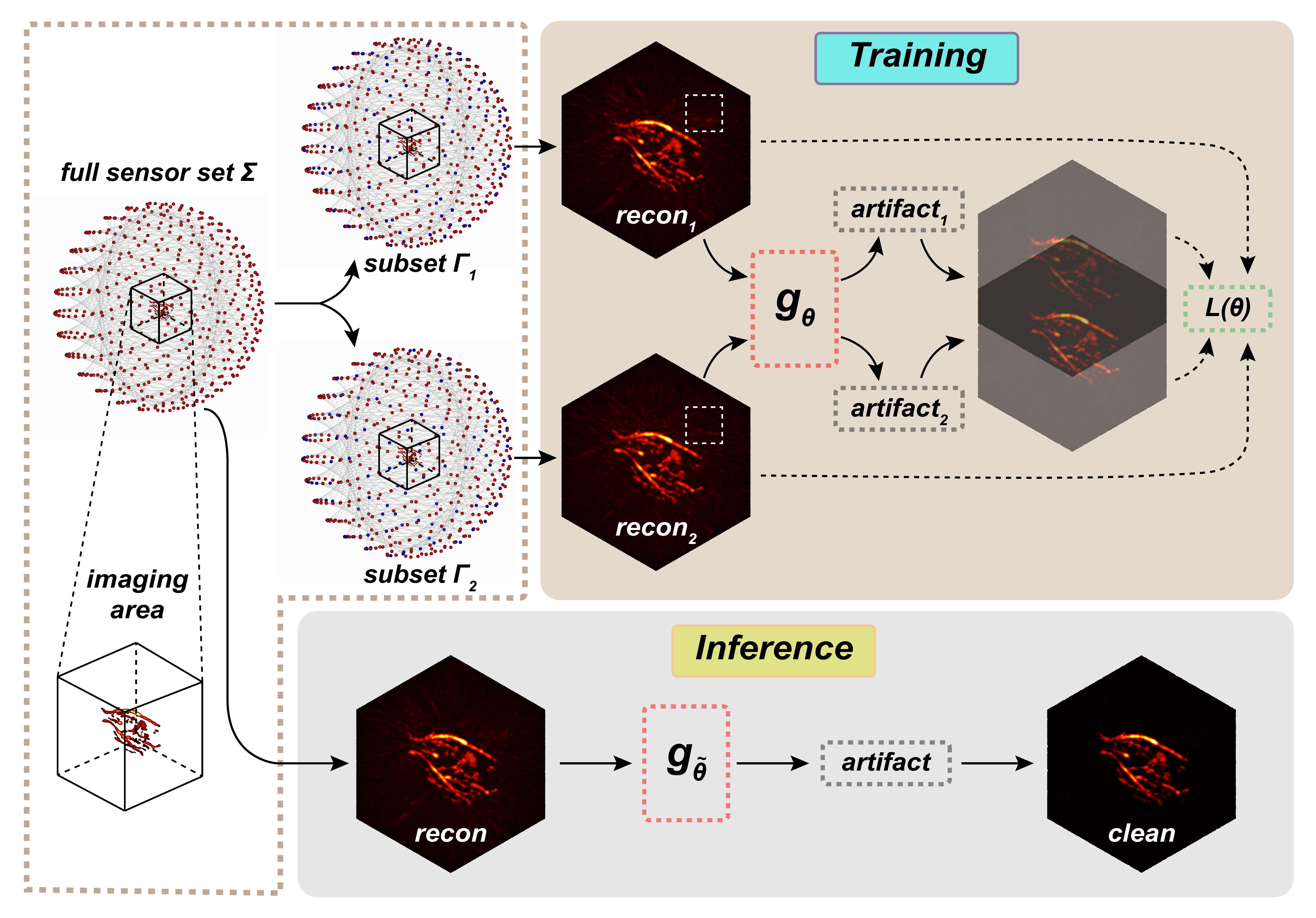}
    \caption{The overview of Zero-Shot Artifact2Artifact pipeline.}
    \label{fig:fig3}
\end{figure}

The workflow of ZS-A2A is illustrated in Fig.~\ref{fig:fig3}. Specifically, for a set of raw PA data $\mathit{\Sigma}$, we apply the randomly discarding strategy to generate two data subsets (Eq.~\ref{eq:eq3}), $\mathit{\Gamma_1}$ and $\mathit{\Gamma_2}$. Next, both subsets, $\mathit{\Gamma_1}$ and $\mathit{\Gamma_2}$, are independently reconstructed using the UBP method, resulting in reconstruction results $\mathit{recon_1}$ and $\mathit{recon_2}$. These reconstruction results are then paired and fed into $\mathit{Latent\ space-Artifact\ predictor}$, a lightweight convolutional neural network~\cite{mansour2023zero} $\mathit{g_\theta}$, to predict the artifacts: $\mathit{artifact}_1 = g_{\theta}(\mathit{recon}_1)$ and $\mathit{artifact}_2 = g_{\theta}(\mathit{recon}_2)$. By minimizing the loss function $\mathit{L(\theta)}$, we optimize the network parameters $\mathit{\widetilde{\theta}}$. Subsequently, we use the optimized network to predict the $\mathit{artifact}$ from the UBP reconstruction result $\mathit{recon}$ of the original PA data $\mathit{\Sigma}$. Finally, by subtracting the predicted $\mathit{artifact}$ from $\mathit{recon}$, we obtain the clean reconstruction result: $\mathit{clean} = \mathit{recon}-g_{\widetilde{\theta}}(\mathit{recon})$.

For ZS-A2A, the loss function $\mathit{L(\theta)}$ is composed of a residual loss $\mathit{L}_\mathrm{res.}(\theta)$ and a consistency loss $\mathit{L}_\mathrm{cons.}(\theta)$. We also use a symmetric loss, which yields the residual loss~\cite{chen2021exploring, mansour2023zero}:

\begin{equation}
    \mathit{L}_\mathrm{res.}(\theta) =\frac{1}{2}\left( \|\mathit{recon}_1 - g_{\theta}(\mathit{recon}_1) - \mathit{recon}_2\|_2^2 + \|\mathit{recon}_2 - g_{\theta}(\mathit{recon}_2) - \mathit{recon}_1\|_2^2 \right).
    \label{eq:eq5}
\end{equation}

Additionally, considering that reconstruction results $\mathit{recon_1}$ and $\mathit{recon_2}$ should converge toward the same high-quality PAI reconstruction result after artifact removal, we impose a consistency loss to serve as a regularization constraint. The consistency loss ensures that the predicted artifact-corrected reconstructions are aligned, encouraging stability and coherence in the reconstruction process. It can be formulated as:  

\begin{equation}
    \mathit{L}_\mathrm{cons.}(\theta) = \| (\mathit{recon}_1 - g_{\theta}(\mathit{recon}_1)) - (\mathit{recon}_2 - g_{\theta}(\mathit{recon}_2)) \|_2^2.
    \label{eq:eq6}
\end{equation}

Therefore, the total loss function is $\mathit{L}(\theta) = \mathit{L}_\mathrm{res.}(\theta) + \mathit{L}_\mathrm{cons.}(\theta)$, and through gradient descent to minimize $\mathit{L}(\theta)$, we get the network parameters $\mathit{\widetilde{\theta}}$ for ZS-A2A.

\section{Experiments and results}

\subsection{Simulation validation of artifact removal for PA image reconstruction.}

We first validated ZS-A2A through simulation studies. In simulation studies, the detector array was configured as a spherical array with a radius of 60 mm, containing 2,048 detector elements arranged across the entire sphere surface following a Fibonacci sequence layout. We used both simple phantom and complex vessel as the simulated PA source, which were located within a cubic region of $x, y, z \in \left(-12.8\text{ mm}, 12.8\text{ mm}\right)$ inside the spherical array (assuming the origin of the coordinate system is located at the center of the spherical array). 

We used the reconstruction results from a full set of 2,048 detectors as the ground truth and uniformly subsampled 512 detectors to simulate PAI reconstruction under sparse detector configurations, resulting in artifact-contaminated reconstruction results to be processed. Subsequently, we applied a random discarding strategy to generate subsets of data. For the simple phantom, the subset contained data from 448 detectors, while for the complex vessel, the subset contained 400 detectors. The reconstruction grid resolution was set to 0.1 mm. 

\begin{figure}[H]
    \centering
    \includegraphics[width=0.85\linewidth]{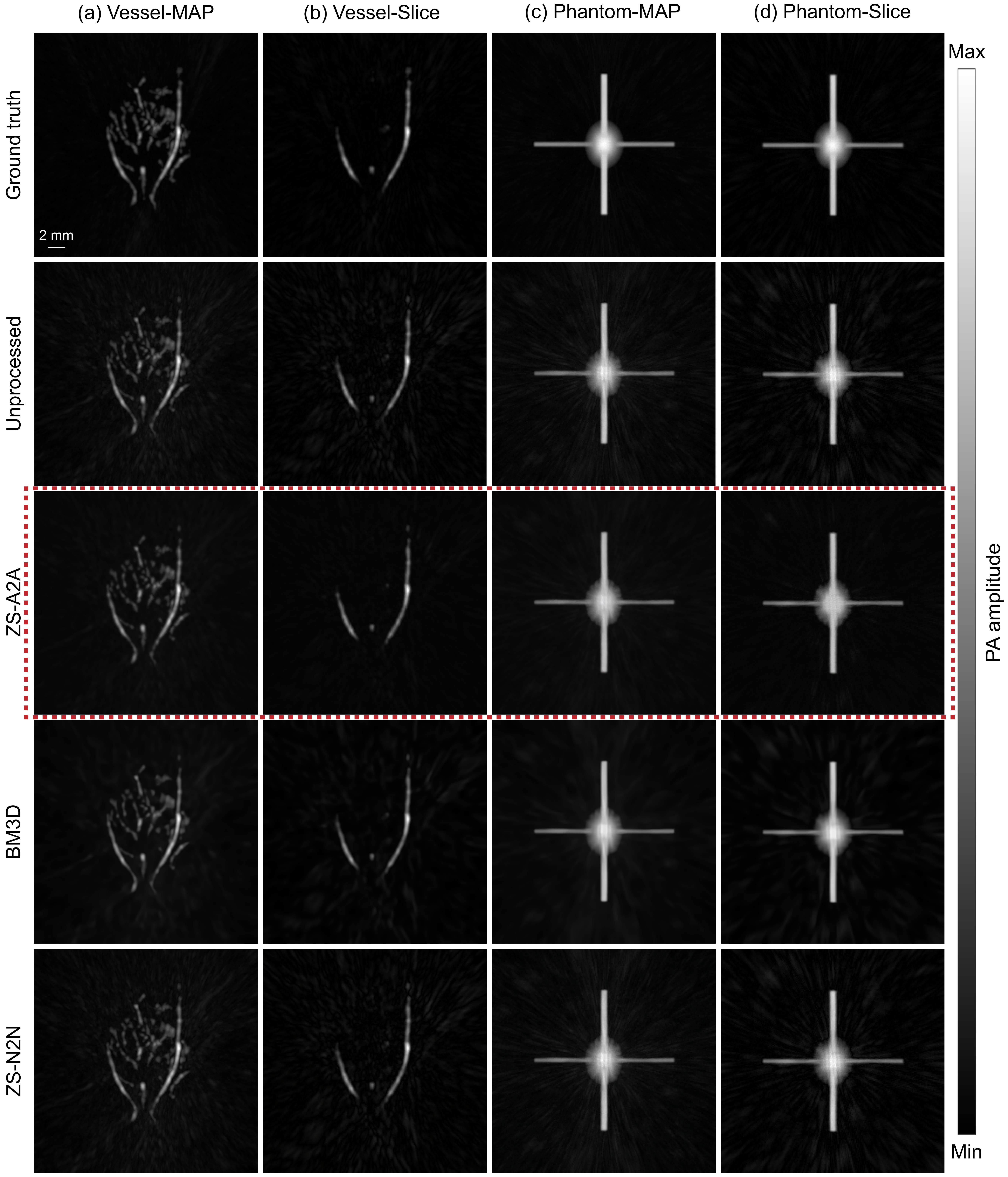}
    \caption{Comparison of results using different zero-shot artifact removal methods. (a) Artifact removal results for image of complex vessel-MAP. (b) Artifact removal results for images of complex vessel-slice. (c) Artifact removal results for image of simple phantom-MAP. (d) Artifact removal results for images of simple phantom-slice. (Scale: 2.0 mm.)}
    \label{fig:fig4}
\end{figure}

In the simulation experiments, ZS-A2A directly performed artifact removal on both slices and maximum amplitude projection (MAP) of the 3D volumes. The results were compared with Zero-Shot Noise2Noise (ZS-N2N), another zero-supervised method, as well as the classical BM3D~\cite{dabov2007image} algorithm. The comparative results are shown in Fig.~\ref{fig:fig4}. To further evaluate the artifact removal performance, Peak Signal-to-Noise Ratio (PSNR) and Contrast-to-Noise Ratio (CNR) metrics were calculated, as displayed in Tab.~\ref{tab:psnr_results} and Tab.~\ref{tab:cnr_results}. It is evident that ZS-A2A achieved state-of-the-art (SOTA) performance across nearly all tasks. 

\begin{table}[ht]
\centering
\caption{PSNR (in dB) of  PAI images}
\label{tab:psnr_results}
\renewcommand{\arraystretch}{1.3} 
\setlength{\tabcolsep}{8pt} 
\begin{tabular}{c|c|c|c|c} 
\hline
\textnormal{Image} & \textbf{Vessel-MAP} & \textbf{Vessel-Slice} & \textbf{Phantom-MAP} & \textbf{Phantom-Slice} \\
\hline
\textnormal{Unprocessed} & 30.59 & 32.66 & 29.91 & 31.11 \\
\textnormal{ZS-A2A} & \textbf{31.22} & \textbf{34.83} & 30.24 & \textbf{32.61} \\
\textnormal{BM3D} & 31.11 & 34.05 & \textbf{30.52} & 32.23 \\
\textnormal{ZS-N2N} & 30.60 & 32.66 & 29.93 & 31.15 \\
\hline
\end{tabular}
\end{table}

\begin{table}[ht]
\centering
\caption{CNR of PAI images}
\label{tab:cnr_results}
\renewcommand{\arraystretch}{1.3} 
\setlength{\tabcolsep}{8pt} 
\begin{tabular}{c|c|c|c|c} 
\hline
\textnormal{Image} & \textbf{Vessel-MAP} & \textbf{Vessel-Slice} & \textbf{Phantom-MAP} & \textbf{Phantom-Slice} \\
\hline
\textnormal{Unprocessed} & 26.18 & 13.32 & 24.61 & 38.61 \\
\textnormal{ZS-A2A} & \textbf{66.89} & \textbf{34.49} & \textbf{55.24} & \textbf{110.04} \\
\textnormal{BM3D} & 49.54 & 18.66 & 34.36 & 82.44 \\
\textnormal{ZS-N2N} & 26.49 & 13.49 & 25.07 & 39.92 \\
\hline
\end{tabular}
\end{table}

For the artifact removal of individual slices from the complex vessel, ZS-A2A improved the PSNR by more than 2 dB compared to the original image (Tab.~\ref{tab:psnr_results}). This is a significant enhancement for images with an original PSNR in the range of 30 dB. It is worth mentioning that, for the simple phantom, the spatial symmetry and the compression of artifact randomness in MAP result in BM3D achieving reasonable artifact removal performance for simple phantoms, and its PSNR for MAP slightly surpasses that of ZS-A2A. However, as the phantom complexity increases and becomes closer to realistic conditions, the performance of BM3D noticeably deteriorates. 

Additionally, from the perspective of CNR (as shown in Tab.~\ref{tab:cnr_results}), ZS-A2A demonstrates an unparalleled ability to enhance image quality. This improvement is also evident in the results displayed in Fig.~\ref{fig:fig4}. The artifact removal effectiveness for the complex vessel signals the potential of ZS-A2A to significantly enhance the quality of real PA images.

\subsection{Artifact removal for 3D PA image reconstruction of in vivo animal studies.}

Then, we performed artifact removal capability of ZS-A2A for the 3D PAI reconstruction results of \( in\ vivo \) rat experimental data. The \( in\ vivo \) animal study data, including rat liver and rat kidney, was acquired by Kim's lab using a hemispherical ultrasound (US) transducer array with 1,024 elements and a radius of 60 mm~\cite{choi2023deep}. Each US transducer element in the array had an average center frequency of 2.02 MHz and a bandwidth of 54\%. The effective field of view (FOV) was $12.8 \text{ mm} \times 12.8 \text{ mm} \times 12.8 \text{ mm}$, and the spatial resolutions of approximately 380 µm were nearly isotropic in all directions when all 1,024 US transducer elements were used~\cite{choi2023deep}. More details regarding the 3D PAI system and animal experiments can be found in literatures~\cite{choi2023deep, kim20243d, kim2022deep}.

\begin{figure}[H]
    \centering
    \includegraphics[width=0.9\linewidth]{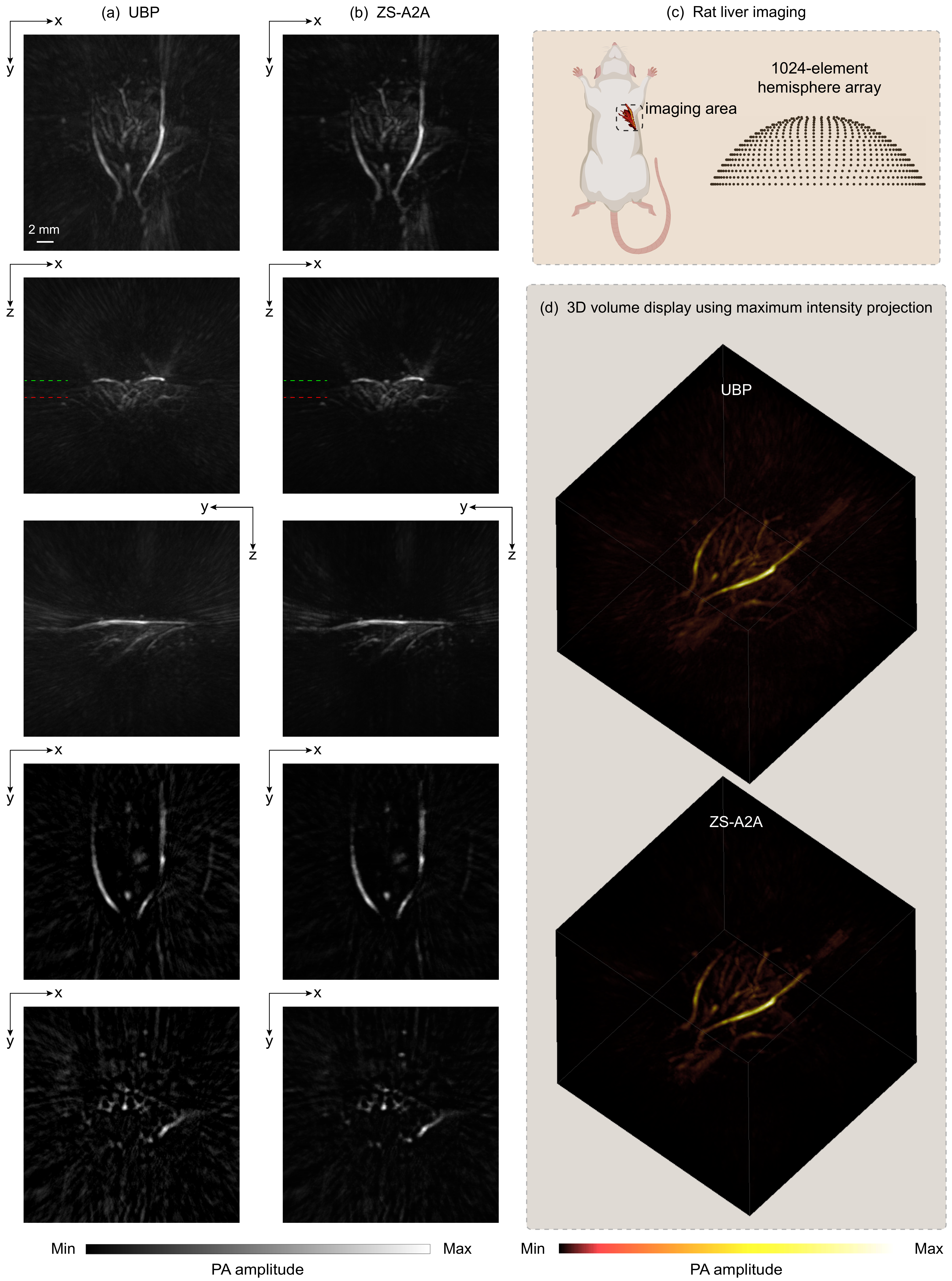}
    \caption{3D PA reconstruction results of a rat liver. (a) XY Plane-MAP, XZ Plane-MAP, YZ Plane-MAP and the cross-section slice at green dashed line and red dashed line marked in XZ Plane-MAP of the UBP 3D reconstruction results using 1,024 sensor signals. (b) XY Plane-MAP, XZ Plane-MAP, YZ Plane-MAP and the cross-section slice at green dashed line 
     and red dashed line marked in XZ Plane-MAP of the UBP 3D reconstruction results after ZS-A2A artifact removal using 1,024 sensor signals. (Scale: 2 mm.) (c) Schematic diagram of the imaging area. (d) 3D volume display of the reconstruction results using maximum intensity projection.}
     \label{fig:fig5}
\end{figure}

\begin{figure}[H]
    \centering
    \includegraphics[width=0.9\linewidth]{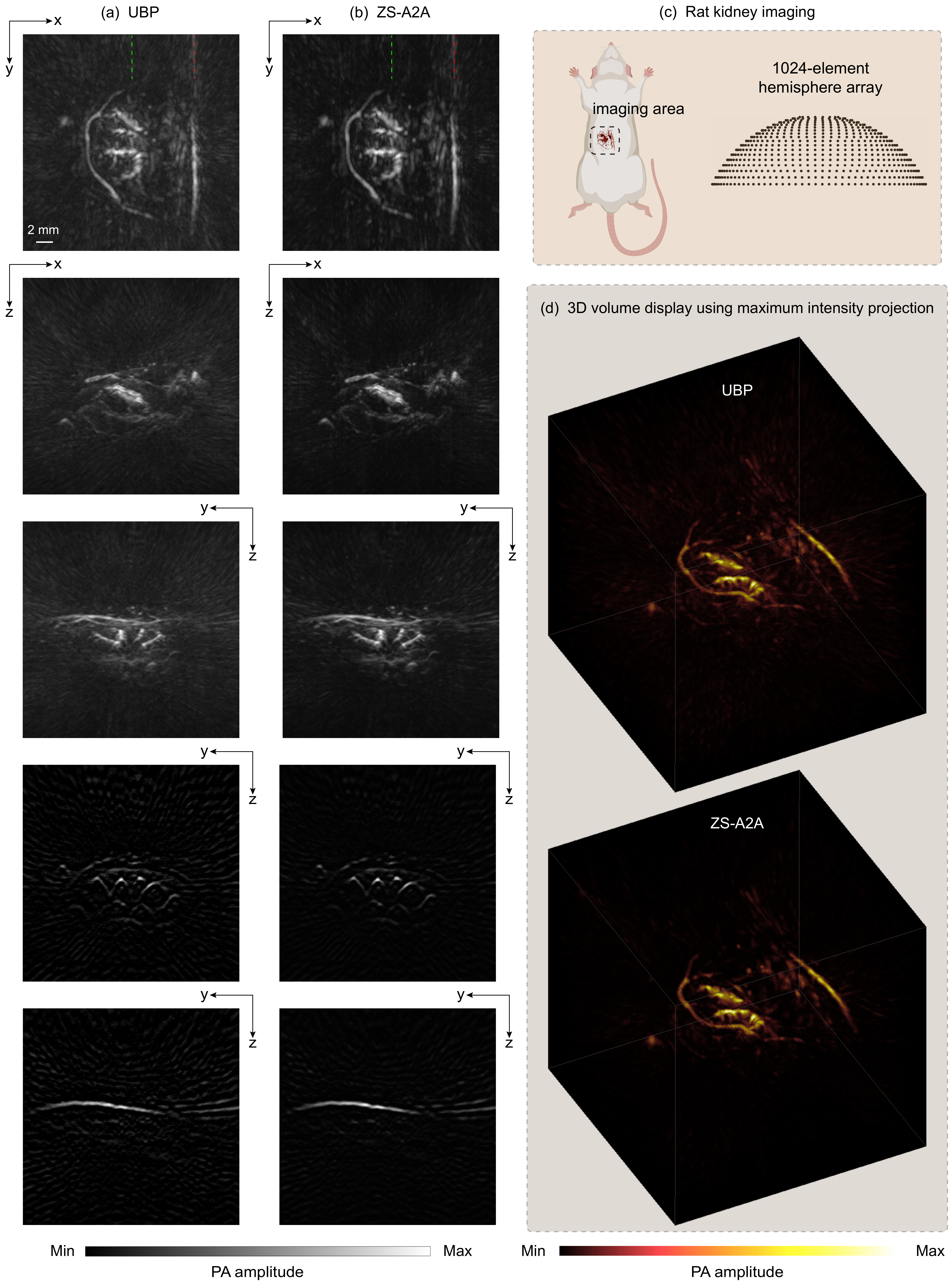}
    \caption{3D PA reconstruction results of a rat kidney. (a) XY Plane-MAP, XZ Plane-MAP, YZ Plane-MAP and the cross-section slice at green dashed line and red dashed line marked in XY Plane-MAP of the UBP 3D reconstruction results using 1,024 sensor signals. (b) XY Plane-MAP, XZ Plane-MAP, YZ Plane-MAP and the cross-section slice at green dashed line 
     and red dashed line marked in XY Plane-MAP of the UBP 3D reconstruction results after ZS-A2A artifact removal using 1,024 sensor signals. (Scale: 2 mm.) (c) Schematic diagram of the imaging area. (d) 3D volume display of the reconstruction results using maximum intensity projection.}
     \label{fig:fig6}
\end{figure}

We generated subsets of data using a random discarding strategy, retaining 800 detectors out of 1,024 total detectors for our experiments. For the 3D PAI reconstruction results, we applied ZS-A2A to each slice independently to obtain the artifact-free 3D reconstruction, without requiring any modifications to the network's training parameters.  

We presented the reconstruction results of a rat liver and rat kidney in Fig.~\ref{fig:fig5} and Fig.~\ref{fig:fig6}, respectively. Fig.~\ref{fig:fig5}(a-b) and Fig.~\ref{fig:fig6}(a-b) (from top to bottom) show the maximum amplitude projection (MAP) images along three orthogonal directions of the original 3D reconstruction results obtained using UBP and the 3D reconstruction results after applying ZS-A2A. The slices corresponding to the green and red dashed lines are also displayed. The results clearly demonstrate ZS-A2A's superior artifact removal ability.  

We further quantified the improvement in image quality using CNR based on the slice results before and after applying ZS-A2A. For the liver, the CNR improved from 17.48 to 43.46, and for the kidney, the CNR increased from 21.56 to 46.31. In addition, we visualized the 3D results (based on maximum intensity projection) before and after artifact removal in Fig.~\ref{fig:fig5}(d) and Fig.~\ref{fig:fig6}(d). The visualizations show a significant enhancement in reconstruction quality, further confirming the effectiveness of ZS-A2A.

\subsection{Computational time.}
 
Computation time is an essential metric for evaluating the efficiency of a model. In our experiments, the number of iterations was set to 3,000, with a learning rate of 0.01, a learning rate decay schedule of \( step\_size = 1,000 \), and \( \gamma = 0.6 \). In the simulation studies, the input of a single slice or maximum amplitude projection (MAP) image with the size of $ 500 \times 500$, takes about 100 seconds to complete both the learning and artifact removal inference on a single NVIDIA GeForce RTX 2060 SUPER. However, on a single NVIDIA GeForce RTX 3090 Ti, for a single slice with the size of $256 \times 256$, corresponding to the reconstructed region of a typical PA hemispherical array imaging system ($25.6 \text{ mm} \times 25.6 \text{ mm} \times 25.6 \text{ mm}$) with a reconstruction grid resolution of 0.1 mm, ZS-A2A required less than 8 seconds to complete both the learning and artifact removal inference, and for the entire 3D reconstruction (as in the \( in\ vivo \) experimental data validation presented in this work), the process took only 25 minutes to complete the entire learning and inference for artifact removal.

\section{Ablation}

In ZS-A2A, we generated only two subsets of data via random detector discarding to learn the artifact patterns. Interestingly, when the number of subsets is increased, the artifact removal performance of the model diminishes. This is likely because the network begins to learn deeper patterns of the artifacts under the randomized perturbations, which prevents it from effectively distinguishing these artifacts from the true PA signals. For the artifact removal experiment on the MAP image of a simulated complex vessel, we increased the number of subsets to 3 and correspondingly modified the calculation of residual and consistency losses. On the same NVIDIA GeForce RTX 2060 SUPER, we observed that, compared to using 2 subsets, the computation time increased from 100 seconds to 120 seconds, while the PSNR dropped from 31.22 to 30.77. When the number of subsets was further increased to 4, the computation time rose to 150 seconds, and the PSNR remained approximately 30.77. This demonstrates that using 2 subsets provides the optimal artifact removal performance for the ZS-A2A framework.

Additionally, like ZS-N2N, ZS-A2A employs a very simple two-layer image-to-image network~\cite{mansour2023zero}. The network consists of two convolution operators with kernel sizes of $3 \times 3$, followed by a $1 \times 1$ convolution operator, resulting in a total of approximately 22k parameters. Unlike ZS-N2N's findings, where increasing network complexity significantly degraded performance (e.g., using U-Net for ZS-N2N led to notable reductions in denoising efficacy~\cite{mansour2023zero}), our experiments revealed that a slight increase in network complexity does not significantly impair artifact removal performance for ZS-A2A. Specifically, we experimented with a relatively shallow U-Net~\cite{ronneberger2015u} (only two down-sampling/up-sampling layers) with approximately 168k parameters. Across multiple tests, the PSNR fluctuated between 31.05 and 31.25, comparable to the lightweight convolutional network’s result of 31.22. However, it is important to note that the shallow U-Net required approximately 430 seconds for learning and inference on the same NVIDIA GeForce RTX 2060 SUPER, which is significantly slower than the lightweight convolutional network (approximately 100 seconds). Therefore, although the shallow U-Net achieves comparable artifact removal performance, its efficiency is far inferior. These results suggest that the current lightweight convolutional network structure used in ZS-A2A is both highly effective and efficient, making it particularly suitable for practical applications.

\section{Discussion}

The proposed Zero-Shot Artifact2Artifact (ZS-A2A) method achieves a significant advancement in the field of photoacoustic imaging (PAI), particularly in addressing the critical issue of reconstruction artifacts arising from sparse and angle-limited detector arrays. Our method leverages the unique properties of PAI artifacts and introduces a novel self-supervised learning paradigm that obviates the need for extensive training datasets or prior knowledge of the artifacts. This innovation is crucial for the practical deployment of PAI in clinical settings, where the variability of imaging systems and the constraints on computational resources are paramount considerations.

The performance of ZS-A2A has been rigorously evaluated through both simulation studies and in vivo animal experiments, demonstrating its superior artifact removal capabilities compared to existing zero-shot methods such as Zero-Shot Noise2Noise (ZS-N2N) and classical BM3D algorithms. In the simulation studies, ZS-A2A achieved state-of-the-art (SOTA) performance across various metrics, including Peak Signal-to-Noise Ratio (PSNR) and Contrast-to-Noise Ratio (CNR). In the in vivo animal studies, ZS-A2A demonstrated its robustness and versatility by effectively removing artifacts from 3D PAI reconstructions of rat liver and kidney data. The quantitative improvements in CNR, from 17.48 to 43.46 for the liver and from 21.56 to 46.31 for the kidney, further validated the method's efficacy. From a computational perspective, ZS-A2A's lightweight network architecture ensures rapid convergence and inference times, making it highly suitable for real-time applications. On a single NVIDIA GeForce RTX 3090 Ti GPU, ZS-A2A could process a single $256 \times 256$ slice in less than 8 seconds and a full 3D volume in approximately 25 minutes. This computational efficiency is a critical advantage over iterative methods, which often require hours to process similar datasets.

However, ZS-A2A still has certain limitations. When the detector array is overly sparse or the angular coverage is severely restricted, ZS-A2A may not be able to effectively eliminate artifacts. Additionally, under the same detector data conditions, the reconstruction quality of ZS-A2A cannot match that achieved by iterative methods. Nevertheless, considering computational efficiency, ZS-A2A demonstrates greater practicality and usability compared to iterative methods.

In conclusion, the development of ZS-A2A represents a significant step forward in improving the quality of PAI reconstructions. Its zero-shot learning capability, combined with computational efficiency and robust performance, makes it a highly promising tool for enhancing the clinical utility of PAI. Furthermore, the ZS-A2A method has the potential to be extended to other imaging modalities such as PET and SPECT, offering a generalized framework for artifact removal in medical image reconstruction. It may serve as a universal solution for enhancing the quality of reconstructed images across a wide range of medical imaging techniques.

\section*{Acknowledgments}

This research was supported by the following grants: the National Key R\&D Program of China (No. 2023YFC2411700, No. 2017YFE0104200); the Beijing Natural Science Foundation (No. 7232177); the Basic Science Research Program through the National Research Foundation of Korea (NRF) funded by the Ministry of Education (2020R1A6A1A03047902).

\bibliographystyle{ieeetr}
\bibliography{references}

\end{document}